\documentclass[runningheads]{llncs}
\usepackage[bookmarks=true, pdftex, colorlinks=true, plainpages=false,
pdfauthor={}, pdftitle={}, pdfkeywords={}]{hyperref}
\usepackage{graphicx}
\usepackage{amsfonts}
\usepackage{multirow}
\usepackage{color}
\DeclareGraphicsExtensions{.jpg,.pdf,.mps,.png,.ps}
\usepackage{booktabs}
\usepackage{microtype}
\usepackage{rotating}
\usepackage{bm}
\usepackage{cite}

\usepackage{amsmath}
\usepackage{bbm}

\usepackage{esvect}
\usepackage{caption}
\usepackage[linesnumbered,ruled,vlined]{algorithm2e}

\usepackage{caption}
\usepackage{algorithmic}
\usepackage[font=footnotesize]{subcaption}
\SetCommentSty{mycommfont}
\SetKwInput{KwInput}{Input}                
\SetKwInput{KwOutput}{Output}              

\begin{document}
\pdfpageattr {/Group << /S /Transparency /I true /CS /DeviceRGB>>}

\title{Self domain adapted network}

\titlerunning{Self domain adapted network}

\author{Yufan He\inst{1} \and Aaron Carass\inst{1}, Lianrui Zuo\inst{1,3}, Blake E. Dewey\inst{1} \and Jerry~L.~Prince\inst{1,2}}
\authorrunning{Y.~He et al.}
\institute{$^{1}$Dept. of Electrical and Computer Engineering,
$^{2}$Dept. of Computer Science,\\
The Johns Hopkins University, Baltimore, MD~21218,~USA\\[0.75em]
$^{3}$Laboratory of Behavioral Neuroscience, National Institute on Aging, \\ National Institute of Health, Baltimore, MD 20892,~USA}



\maketitle

\begin{abstract}
Domain shift is a major problem for deploying deep networks in  clinical practice. Network performance drops significantly with (target) images obtained differently than its (source) training data. Due to a lack of target label data, most work has focused on unsupervised domain adaptation~(UDA). Current UDA methods need both source and target data to train models which perform image translation~(harmonization) or learn domain-invariant features. However, training a model for each target domain is time consuming and computationally expensive, even infeasible when target domain data are scarce or source data are unavailable due to data privacy. In this paper, we propose a novel self domain adapted network~(SDA-Net) that can rapidly adapt itself to a single test subject at the testing stage, without using extra data or training a UDA model. The SDA-Net consists of three parts: adaptors, task model, and auto-encoders. The latter two are pre-trained offline on labeled source images. The task model performs tasks like synthesis, segmentation, or classification, which may suffer from the domain shift problem. At the testing stage, the adaptors are trained to transform the input test image and features to reduce the domain shift as measured by the auto-encoders, and thus perform domain adaptation. We validated our method on retinal layer segmentation from different OCT scanners and T1 to T2 synthesis with T1 from different MRI scanners and with different imaging parameters. Results show that our SDA-Net, with a single test subject and a short amount of time for self adaptation at the testing stage, can achieve significant improvements.
\end{abstract}

\section{Introduction}
\label{s:intro}
The success of deep networks relies on the assumption that test data~(target) and training data~(source) are generated from the same distribution. In real scenarios such as multi-center longitudinal studies---even with a pre-defined imaging protocol---subjects will occasionally be scanned by different scanners with different imaging parameters. The model that is trained on the source data will incur a significant performance drop on those scans, which can vary from the source domain in different ways~(the domain shift problem), and each target domain may only contain a few subjects. 

Unsupervised domain adaptation~(UDA), which reduces the domain shift without any target labels, is often used for solving domain shift problems. UDA can be categorized into two types. The first type is data harmonization in the pixel domain, which translates the target image to be similar to the source image. This includes methods like histogram matching, style transfer~\cite{gatys2016image}, and  Cycle-Gan~\cite{zhu2017unpaired}.  Ma et al.~\cite{ma2019neural} used style transfer~\cite{gatys2016image} to reduce the effect of domain shift in MR heart image segmentation.  Seeböck et al.~\cite{seebock2019using} used a Cycle-Gan to improve OCT lesion segmentation. The second type of UDA learns domain-invariant features. In particular, the network is re-trained to produce domain-invariant features from the source and target data such that those features are similar, as measured by metrics like maximum mean discrepancy~\cite{long2015learning}, or are indistinguishable by domain classifiers~\cite{ganin15pmlr} or discriminators in  adversarial training~\cite{tzeng2017adversarial,tsai2018learning,dou2018unsupervised}. A combination of these two approaches is proposed in~\cite{hoffman2018cycada}.
However, those methods~(except histogram matching and style transfer)  1)~require retraining a UDA model for each target domain which is time consuming and computationally expensive; 2)~require  a fair amount of target data from each domain to train the model, which may not be feasible in clinical practice; or 3)~require source data which may not be available for people deploying a pre-trained model due to data privacy. Some work~\cite{ouyang2019data,benaim2018one} has addressed problem (2) but no method has addressed all three problems.

Can we design a model that can be rapidly adapted to a single test subject during inference without using extra data? If so, we can directly deploy the trained model on images from various target domains without accessing the source data or retraining a UDA model. In this paper, we propose a new deep model for this purpose and name it the self domain adapted network~(SDA-Net). Adapting a pre-trained classifier to test images was first proposed in~\cite{jain2011online} for multi-face detection, and Assaf et al.~\cite{shocher2018zero} trained a super-resolution model on a single test image. However, those methods cannot be easily modified for our task. The SDA-Net consists of three parts: 1)~a task network~(T) which performs our task~(synthesis, segmentation, or classification); 2)~a set of auto-encoders~(AEs), which are used as alignment measurements; and
3)~a set of adaptors which perform domain adaptation on each test subject during inference. The core idea is to align the target and source domain in the pixel, network feature, and network output~\cite{tsai2018learning} spaces. We have two major differences with the previous methods: 1)~only one target test subject is used for training the adaptors at testing stage, while T and AEs with high training cost are frozen; and 2)~the alignment is measured by the AEs. AEs have been used for anomaly detection under a core assumption: abnormal inputs will have larger reconstruction error than normal inputs~\cite{gong2019memorizing}. We extend this and define the source domain as normal and the target domain as abnormal and use the reconstruction error as an alignment measurement. The adaptors perform domain adaptation by minimizing the AEs' reconstruction error on the target data. 

Implementation of this general framework faces several obstacles. Firstly, AEs have a strong generalization ability such that abnormal inputs with low reconstruction error can be far away from the source features\cite{gong2019memorizing}. Secondly, features from different classes can collapse to one by the adaptors, as illustrated in Fig.~\ref{f:ae}. Thirdly, deep networks can hallucinate features~\cite{cohen2018distribution}, which is a severe problem for medical data. We avoid the first problem by focusing on tasks with relatively minor domain shifts like scanner differences; thus, we can assume that initial target features are close to the source. With this assumption and with specially designed adaptors and training loss, we can successfully address the other problems. 
\label{s:intro}
\section{Method}
\begin{figure}
    \centering
    \includegraphics[width=1\textwidth]{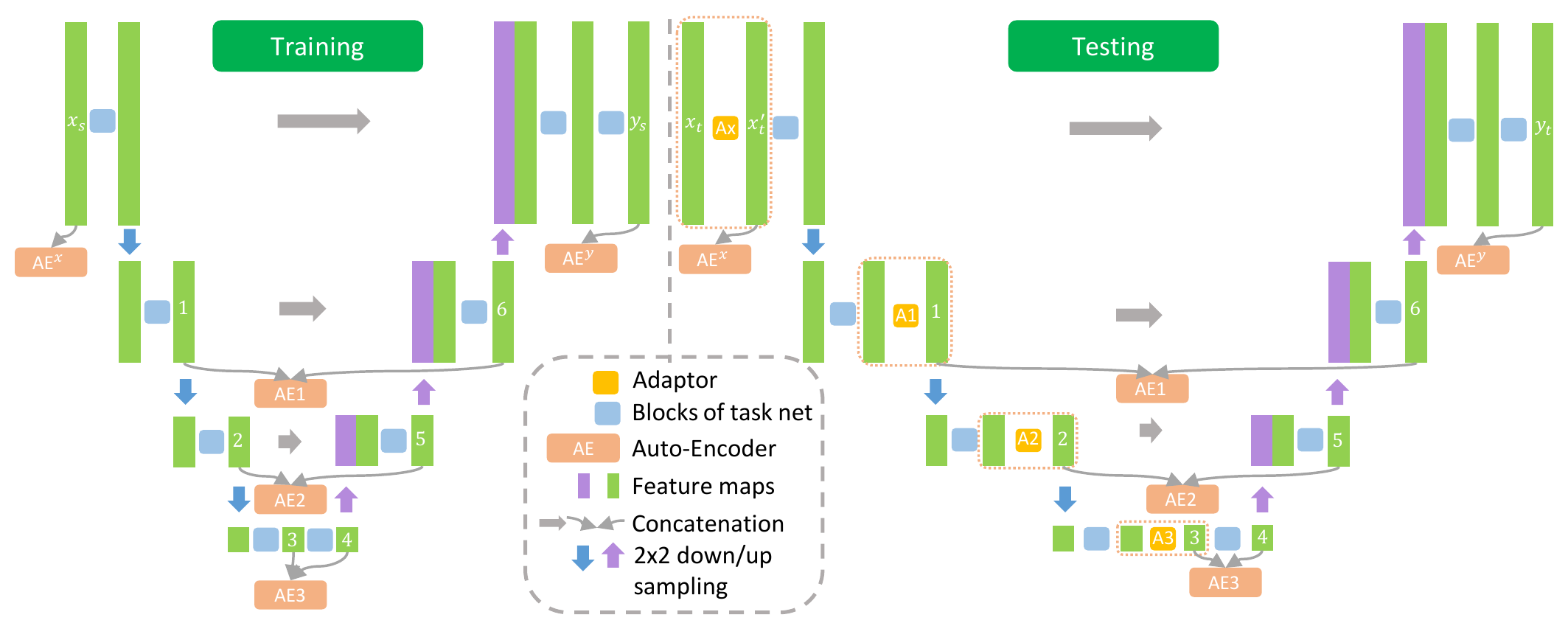}
    \caption{During source domain training with ($x_s, y_s$)~(left figure), the adaptors are not used. We first train the task network then freeze the weights and train AEs. During inference~(right figure), the adaptors transform input target domain image $x_t$ and intermediate features to minimize AE's reconstruction loss.}
    \label{f:framework}
\end{figure}
\begin{figure}[t]
    \centering
    \includegraphics[width=1\textwidth]{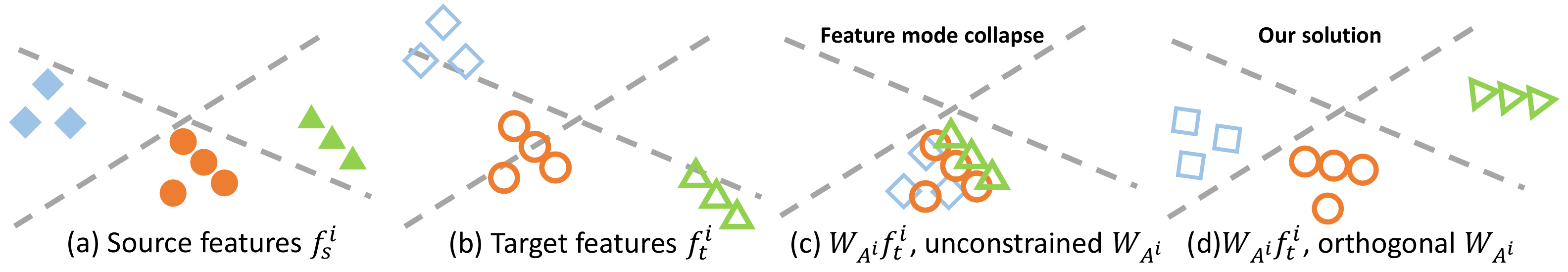}
    \caption{(a) A trained task prediction model~(dotted line) separates source features $f_s^i$ into different classes~(solid points with different colors) (b) Target features $f_t^i$~(hollow points) have domain shifts. (c) $f_t^i$ is transformed by a linear transformation $W_{A^i}$ without constraints, which causes feature mode collapse, and is avoided by imposing orthogonality constraints on $W_{A^i}$ in (d). }
    \label{f:ae}
\end{figure}
\noindent\textbf{Task Network}
We consider two tasks in this paper: retinal OCT layer segmentation and T1 to T2 MRI image synthesis. We do not focus on designing the best task network for these tasks, however, since this is not the main focus of our paper.  Instead, we simply use a residual U-Net~\cite{he2019fully}, a variation of the widely used U-Net~\cite{ronneberger2015miccai}, as our task network for both tasks~(with the only difference being the output channels and output activation). The network has three $2\times2$ max-pooling and 64 channels for all intermediate features as shown in Fig.~\ref{f:framework}. We can replace the task network with any specific state-of-the-art structure.

\noindent\textbf{Multi-level auto-encoders}
A set of fully convolutional auto-encoders~(AEs), $\{AE^x,\{AE^i\}_{i=1,2,3},AE^y\}$ are used.
$AE^x$ and $AE^y$ are trained to reconstruct the source data $x_s$ and T's output $y_s'$, respectively, to encode information in the highest resolution level. As shown in Fig.~\ref{f:framework}, we use three AEs~${\{AE^i\}}_{i=1,2,3}$ to encode T's intermediate features $\{f_s^i\}_{i=1,\cdots,6}$~(all 64 channels) at the lower three resolution levels. ${\{AE^i\}}$'s input is the concatenation of the task network feature $f_s^{i}$ and $f_s^{7-i}$ with the same spatial resolution at level $i$. The AEs' network structure is a modification of T, where two $2\times2$ max-pooling and instance normalization are used, while the long skip connections are removed. The encoder feature channel numbers are 64, 32, 16~(inverse for the decoder) for ${\{AE^i\}}$'s and 32, 16, 8 for $AE^x$ and $AE^y$~($AE^y$'s input channel number equals $y_s'$'s). 

\noindent\textbf{Training}
The task network T is first trained with pairs of source images $x_s$ and labels $y_s$ under standard training procedure by updating network weights $W_T$ to minimize prediction error~(we use cross entropy for segmentation and mean squared error for synthesis). Then we freeze the task network weights $W_T$, and train the AEs to minimize the reconstruction error $\mathcal{L}_{AE}$.
\begin{align*}
    & y_s', \{f_s^i\}_{i=1,\cdots,6}=T(x_s;W_T) \\
    \mathcal{L}_{AE} =& \sum_{i=1}^3|AE^i(\{f_s^{i},f_s^{7-i}\}) - \{f_s^{i},f_s^{7-i}\}|^2_2 \\
    & + |AE^x(x_s) - x_s|^2_2 + |AE^y(y_s') - y_s'|^2_2    
\end{align*}

\noindent\textbf{Domain adaptor}
The adaptors consist of an image adaptor $A^x$ which transforms the input target image $x_t$ in the pixel-domain and three feature adaptors $\{A^i\}_{i=1,2,3}$ which transform the intermediate features $\{f_t^i\}_{i=1,2,3}$ from T in the feature domain~($A^x$ also influences $f_t^i$ by transforming $x_t$). To make the adaptors trainable by a single subject and to prevent hallucination, we limit the transformation ability of the adaptors. The image adaptor $A^x$ is a pure histogram manipulator with three convolutional layers, where each layer has a $1\times1$ convolution followed by leaky ReLU and instance normalization. The output channels of each layer are 64, 64, 1. Each feature adaptor $A^i$ is a $1\times1$ convolution with 64 input and output channels and the weight $W_{A^i}$ is a $64\times64$ linear transformation matrix. For a 64-channel feature map, each pixel has a feature vector of length 64. Consider two 1D feature vectors $f_a$, $f_b$ and their transformation $W_{A^i}f_a$, $W_{A^i}f_b$, we prevent the feature mode collapse~(illustrated in Fig.~\ref{f:ae}) by keeping the $L_2$ distance between them $(f_a-f_b)^TW_{A^i}^TW_{A^i}(f_a-f_b) = (f_a - f_b)^T(f_a - f_b)$
, which requires orthogonality of $W_{A^i}$ such that $W_{A^i}^TW_{A^i}=I$. We impose orthogonality on $W_{A^i}$ by using the Spectral Restricted Isometry Property Regularization~\cite{bansal2018can}, which minimizes the spectral norm of $W_{A^i}^TW_{A^i}-I$. We define it as the orthogonal loss $ \mathcal{L}_{orth}$ in Eqn.~\ref{eq:orth}, and the implementation details are in~\cite{bansal2018can}. We train the adaptors with $\mathcal{L}_{A}=\mathcal{L}_{AE}+\lambda_{orth}\mathcal{L}_{orth}$. The overall algorithm in the testing stage is described in Alg.~\ref{a:alg}.
\begin{gather}
    \mathcal{L}_{orth}=\sum_{i=1}^3\sigma(W_{A^i}^TW_{A^i} - I) =  \sum_{i=1}^3\underset{z\in R^n, z\neq0}{sup}\biggl\lvert\frac{||W_{A^i}z||^2}{||z||^2} - 1\biggr\rvert
    \label{eq:orth}    
\end{gather}

\begin{algorithm}[!tbh]

\SetAlgoLined
\caption{Self domain adapted network}
\label{a:alg}
\KwInput{Single test subject scan $x_t$}
 Load pre-trained task network $T(\cdot;W_T)$ and $\{AE^i\}_{i=1,2,3}, AE^x, AE^y$\;
 Configure learning rate $\eta$\ and loss weight $\lambda_{orth}$ \;
 Initialize $A^x$ with Kaiming normal~\cite{he2015delving} and 
 $\{A^i\}_{i=1,2,3}$ with identity\;
 \While{$\mathrm{iter}<5$ and $ 0.95*\mathcal{L}_{A}^{iter-1}>\mathcal{L}_{A}^{iter}$}{
  Obtain adapted image, adapted features and prediction: $x_t', \{f_t^i\}_{i=1,\cdots,6},y_t'$ from $T(x_t, A^x, \{A^i\}_{i=1,2,3};W_A,W_T)$\;
  Calculate AE reconstruction loss $\mathcal{L}_{AE} = \sum_{i=1}^3|AE^i(\{f_t^{i},f_t^{7-i}\}) - \{f_t^{i},f_t^{7-i}\}|_2^2+ |AE^x(x_t') - x_t'|_2^2 + |AE^y(y_t') - y_t'|_2^2$\;
  Calculate orthogonality loss $\mathcal{L}_{orth} = \sum_{i=1}^3 \sigma(W_{A^i}^TW_{A^i}-I)$\;
  Update  $A^x,\{A^i\}_{\{i=1,2,3\}}$'s weights $W_A = W_A - \eta\nabla_{W_A}\mathcal{L}_{A}$ \;
 }
\KwOutput{Target prediction $y_t'$ from $T(x_t, A^x, \{A^i\}_{i=1,2,3};W_A,W_T)$}
\end{algorithm}

\section{Experiments}
We validated our SDA-Net on two tasks: retinal layer segmentation~\cite{he2019fully,he2019deep} and T1 to T2 synthesis~\cite{jog2015mr,dar2019image}. The hyper-parameters for both tasks are the same~(except $\lambda_{orth}=1,5$ for segmentation and synthesis respectively). The task network, AEs, and adaptors were trained with the Adam optimizer with a learning rate 0.001, batch size 2, and no augmentation. The task network training was stopped based on the source validation set and the AEs were trained for 20 epochs, both using the source training set. The adaptors test time training is in Alg.~\ref{a:alg}. 

\begin{figure}[t]
    \centering
    \includegraphics[width=1\textwidth]{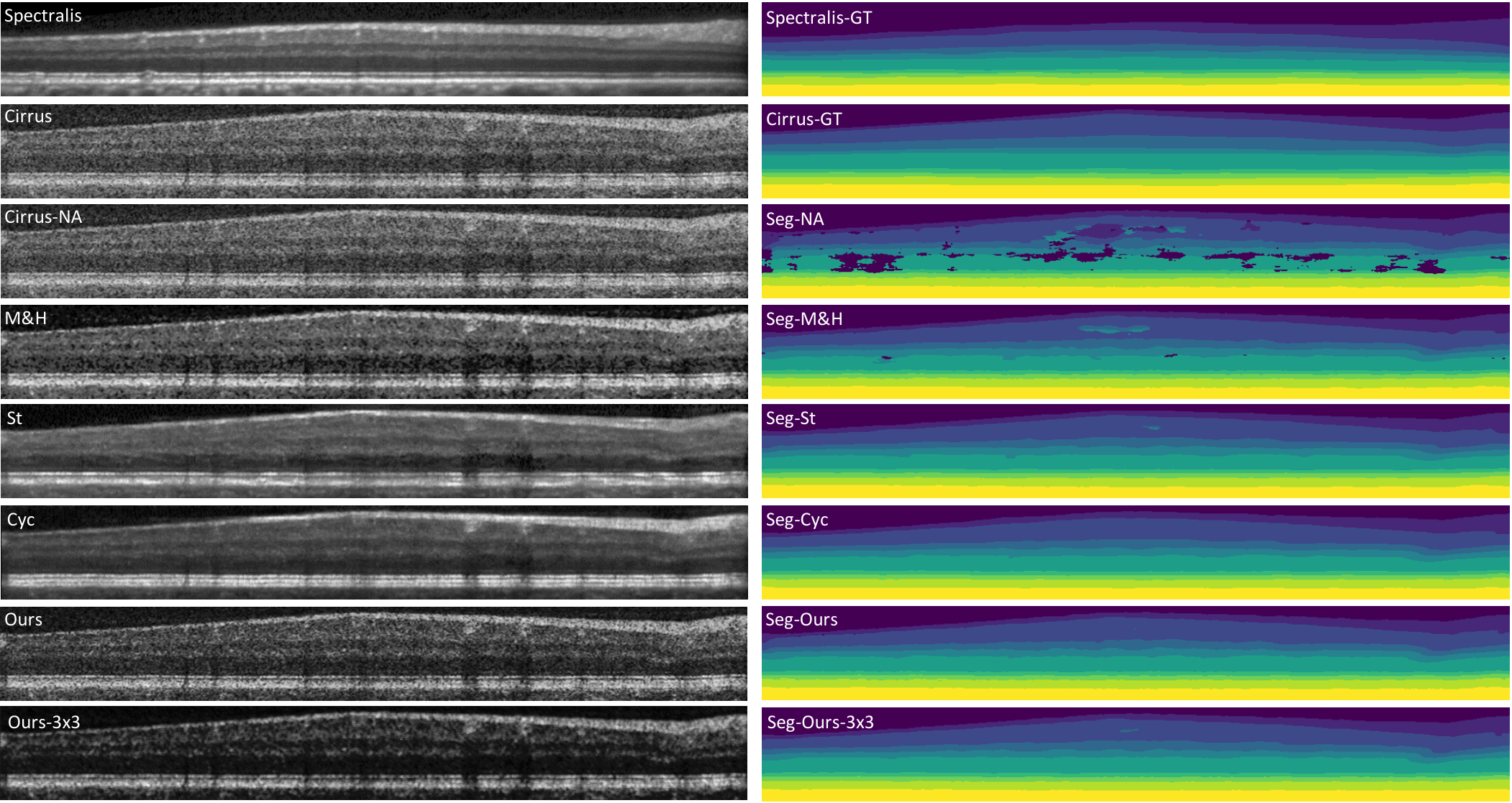}
    \caption{Visualization of source, target and harmonized target images~(Cirrus NA is the original Cirrus) using baselines and our methods~(left). The ground-truth~(GT) and segmentation~(Seg) are shown on the right.}
    \label{f:seg_r}
\end{figure}

\noindent\textbf{Retinal layer segmentation in OCT}
We used retinal images from two OCT scanners: Spectralis and Cirrus. Eight retinal layers were manually segmented and the images were pre-processed with retina flattening~\cite{he2019fully}. Spectralis 2D images~\cite{he2018retinal} were used as the source dataset for training SDA-Net~(588 train, 147 validation, 980 test) and Cirrus images were used as testing target dataset~(6 subjects, each with 8 images). We used SDA-Net to segment each target subject independently. We compare to image harmonization methods without retraining the task network: 1)~NA: No adaptation; 2)~M\&H: $3\times3$ median filter and histogram matching~\cite{seebock2019using}; 3)~St\footnote{\url{https://pytorch.org/tutorials/advanced/neural_style_tutorial.html}}: Style transfer using pre-trained vgg19~\cite{gatys2016image,ma2019neural}; 4)~Cyc\footnote{\url{https://github.com/junyanz/pytorch-CycleGAN-and-pix2pix}}: Cycle-Gan~\cite{zhu2017unpaired}; The Cycle-Gan trained from a single Cirrus subject is not usable. Thus, we trained the Cycle-Gan with 588 Cirrus~(48 test Cirrus images and an additional 540 Cirrus images) and the source training set. We used one image from source training set as the reference image for (2) and (3). We can further improve our results by simply changing the first $1\times1$ convolution of the image adaptor $A^x$ to $3\times3$ for pixel-domain noise removal. We show this result as Ours-$3\times 3$. We also tested SDA-Net on the source testing subject~(20 subjects each with 49 slices). The Dice scores of eight layers and qualitative results are shown in Table~\ref{t:cDice}  and Fig.~\ref{f:seg_r}. The results show that our method improves the target domain results while not significantly affecting the source domain results.
\begin{table}[t]
\caption{Dice scores~(Std. Dev.) for each method. The left part compares domain adaptation performance while the right part illustrates the performance of task model~(NA) and our method on source testing data.}
\label{t:cDice}
\centering

\resizebox{\columnwidth}{!}{%
\begin{tabular}{lcccccc|cc}
\toprule
\multicolumn{7}{c}{Target test results} & \multicolumn{2}{|c}{Source test results}  \\
\hline
\textbf{Layer} &\textbf{NA} &\textbf{M\&H} &\textbf{St} &\textbf{Cyc} &\textbf{Ours} &\textbf{Ours-$3\times3$} &\textbf{Ours} &\textbf{NA}  \\
\hline
\textbf{RNFL}    & 0.615(0.177) & 0.688(0.143) & \textbf{0.724(0.109)} & 0.709(0.126) & 0.682(0.149) & 0.698(0.132) & 0.903(0.040) & 0.906(0.039) \\
\textbf{GCIP}    & 0.742(0.092) & 0.821(0.050) & \textbf{0.841(0.044)} & 0.825(0.042) & 0.818(0.055) & 0.837(0.041) & 0.923(0.032) & 0.927(0.031) \\
\textbf{INL}     & 0.715(0.041) & 0.753(0.033) & 0.767(0.052) & 0.769(0.032) & 0.759(0.029) & \textbf{0.773(0.034)} & 0.822(0.043) & 0.829(0.042) \\
\textbf{OPL}     & 0.612(0.063) & 0.632(0.057) & 0.644(0.059) & 0.670(0.050) & 0.671(0.051) & \textbf{0.704(0.047)} & 0.854(0.029) & 0.856(0.028) \\
\textbf{ONL}     & 0.845(0.031) & 0.859(0.022) & 0.866(0.026) & 0.878(0.026) & 0.892(0.020) & \textbf{0.914(0.018)} & 0.925(0.021) & 0.927(0.023) \\
\textbf{IS}      & 0.803(0.035) & 0.814(0.022) & 0.835(0.024) & 0.811(0.034) & 0.830(0.018) & \textbf{0.838(0.019)} & 0.822(0.033) & 0.818(0.041) \\
\textbf{OS}      & 0.841(0.026) & 0.846(0.028) & 0.833(0.047) & 0.849(0.026) & 0.855(0.024) & \textbf{0.856(0.028)} & 0.839(0.034) & 0.836(0.039) \\
\textbf{RPE}     & 0.820(0.034) & 0.828(0.032) & 0.811(0.042) & \textbf{0.837(0.038)} & 0.834(0.032) & 0.825(0.035) & 0.892(0.040) & 0.890(0.040) \\
\hline
\textbf{Overall} & 0.749(0.089) & 0.780(0.076) & 0.790(0.070) & 0.794(0.067) & 0.793(0.076) & \textbf{0.806(0.070)} & 0.873(0.041) & 0.874(0.042) \\
\bottomrule
\end{tabular}
}
\end{table}
\begin{table}[t]
\centering
\caption{MSE and SSIM~(Std. Dev.) for synthesized T2 evaluated on four datasets~(best result is in bold for each row). HH, GH, and IOP compares domain adaptation performance while Source illustrates the performance of task model~(NA) and our method on source testing data. }
\label{t:syn}
\resizebox{0.9\columnwidth}{!}{%
\begin{tabular}{llccccc}
\toprule
&& {NA} &  {Hist} &  {St} &  {Cyc} &  {Ours} \\ \hline 
\multirow{4}{*}{MSE}  & HH     & 0.223(0.034)           & 0.193(0.049)             & 0.292(0.264)           & 0.186(0.042)            & \textbf{0.168(0.040)}    \\        
                      & GH    & 0.271(0.041)           & 0.237(0.077)             & 0.301(0.204)           & 0.240(0.061)            & \textbf{0.233(0.055)}             \\        
                      & IOP    & 0.329(0.053)           & 0.286(0.064)             & 0.398(0.158)           & 0.297(0.072)            & \textbf{0.279(0.046)}    \\        
                      & Source & 0.092(0.047)           & -                        & -                      & -                       & 0.098(0.047)             \\ \hline 
\multirow{4}{*}{SSIM} & HH     & 0.605(0.051)           & 0.683(0.053)             & 0.575(0.170)           & \textbf{0.700(0.045)}   & 0.693(0.039)             \\        
                      & GH    & 0.595(0.045)           & \textbf{0.671(0.060)}    & 0.601(0.134)           & 0.656(0.041)            & 0.658(0.043)             \\        
                      & IOP    & 0.493(0.067)           & 0.594(0.067)             & 0.472(0.099)           & \textbf{0.620(0.060)}   & 0.564(0.056)             \\        
                      & Source & 0.774(0.050)           & -                        & -                      & -                       & 0.768(0.048)             \\        
\bottomrule
\end{tabular}
}
\end{table}
\begin{figure}[t]
    \centering
    \includegraphics[width=0.85\textwidth]{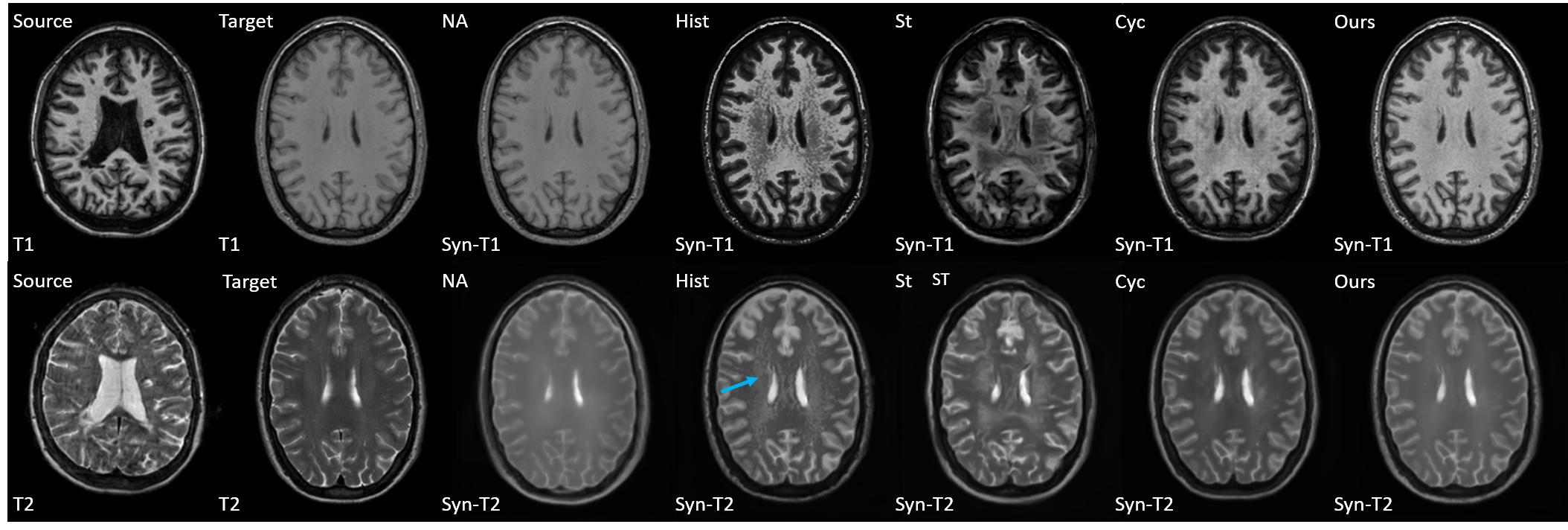}\\
    \includegraphics[width=0.85\textwidth]{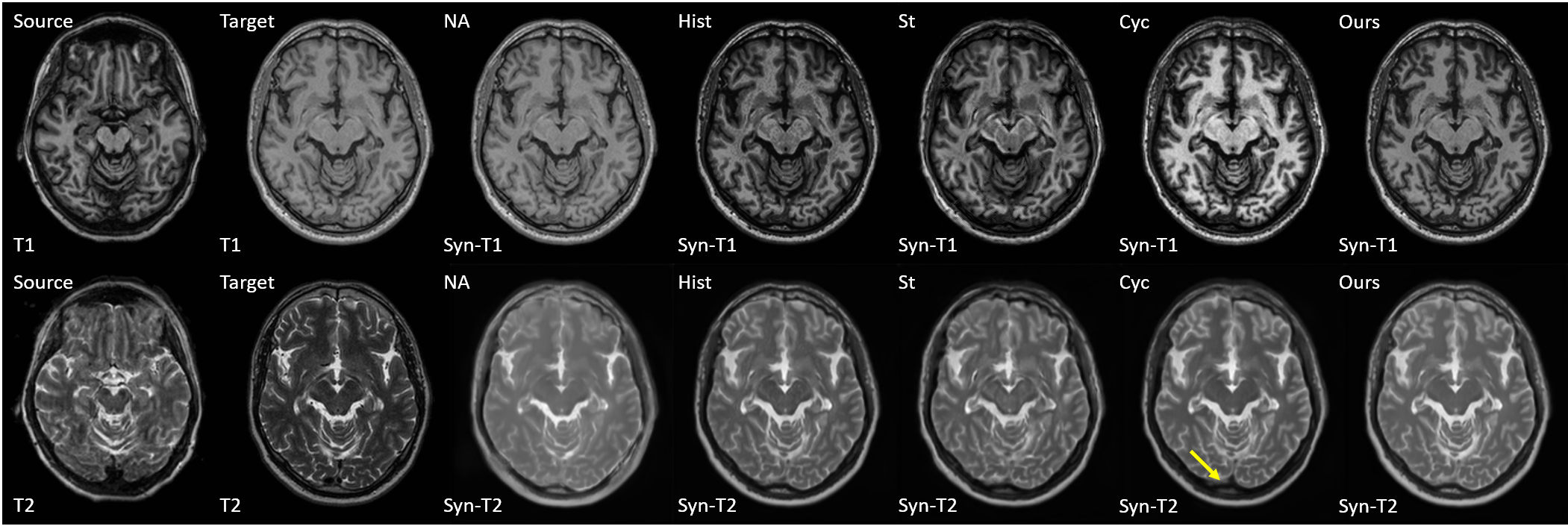} \\
    \includegraphics[width=0.85\textwidth]{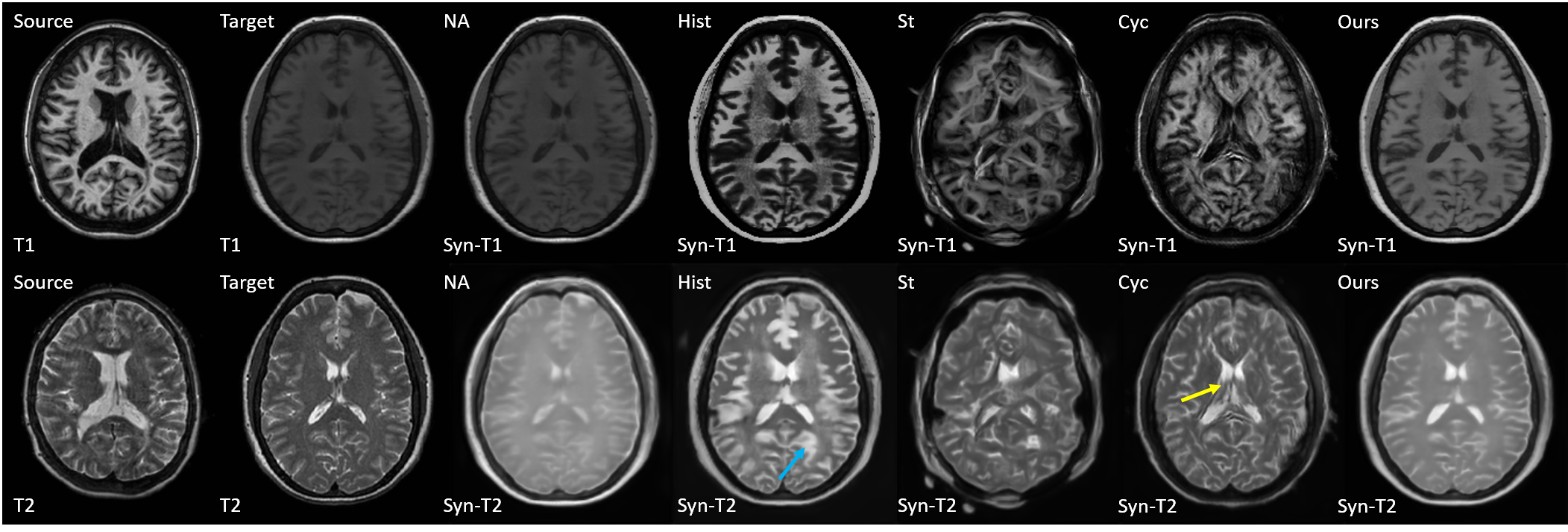}
    \caption{Visualization of harmonized T1 and synthesized T2 from HH~(top two rows), GH~(middle two rows) and IOP~(bottom two rows). The first~(second) column are paired source~(target) T1 and T2. The yellow arrows show a geometry shift and feature hallucination from Cycle-Gan. The blue arrows show artefacts from histogram matching.}
    \label{f:syn_r}
\end{figure}

\noindent\textbf{T1 to T2 synthesis in MRI} 
In the synthesis experiments, we used four datasets which have paired T1-T2 scans. All the scans were N4 corrected~\cite{tustison2010n4itk}, registered to MNI space, and white matter peak normalized. 21 axial slices were extracted from each subject~(equally extracted from slice number 60 to 120, 3mm slice distance). Source dataset (T1 MPRAGE from Philips Achieva 3T,$\text{T}_\text{R}$=3000ms, $\text{T}_\text{E}$=6ms) were used for training the SDA-Net~(630 train, 84 validate, 315 test slices). The target test set comes from the IXI dataset~(https://brain-development.org/ixi-dataset/), which has scans from three different clinical centers~(HH, GH, IOP). We used the first 30 subjects~(630 slices) from HH~(T1 SPGR from Philips Intera 3T,$\text{T}_\text{R}$=9.6ms, $\text{T}_\text{E}$=4.6ms), the first 30 from IOP~(GE 1.5T, unknown parameters) and the first 30 from GH~(T1 SPGR from Philips Gyroscan Intera 1.5T,$\text{T}_\text{R}$=9.8ms, $\text{T}_\text{E}$=4.6ms) for testing T1. As in the OCT segmentation task, we compare to baselines: 1) NA; 2) Hist: 2D histogram matching; 3) St; 4) Cyc: We train three Cycle-Gans for each clinical center, using all source training slices and test slices from each center separately. The reference image for (2) and (3) is the $i$-th slice of the first source training subject, where $i$ is the slice number~(1 to 21) of the input slice. We calculated the MSE and SSIM on the synthesized T2 from target~(30 subjects for each target domain) and source test sets~(15 subjects). The results are shown in Table~\ref{t:syn} and Fig.~\ref{f:syn_r}.

\section{Discussion and Conclusion}
The SDA-Net adapts itself to a single subject for about 30s~(testing time is 5s without adaptation on the same Nvidia GPU) for both tasks and shows comparable results with Cycle-Gan, which requires extra target data and off-line training. As shown in Fig.~\ref{f:syn_r}, vanilla histogram matching can produce artefacts. Style-transfer can produce artistic results since the content and style are not completely disentangled~\cite{gatys2016image}. The Cycle-Gan can cause geometry shift and hallucinate features~\cite{cohen2018distribution}. A complicated pixel-domain transformation needs the model to extract high level features, which needs a fair amount of training data and labels. In order to train on a few images and avoid geometry shift and hallucination, we re-use those high level features extracted by the task network and transform them with $A^i$, while keeping the pixel-domain adaptor  $A^x$ as simple as possible: only $1\times1$ convolutions~(Ours) or with a single $3\times3$ kernel~(Ours-$3\times3$). Despite the simplicity of $A^x$, SDA-Net achieves significant segmentation and synthesis improvements~(Ours) as shown in Table~\ref{t:cDice} and Table~\ref{t:syn}. For segmentation, Ours-$3\times3$ shows that a task specific adaptor can further improve the results. The major limitation of the work is that we only focus on problems with minor domain shift where we assume the features from task network can be re-used and not far from source features. Although we are not solving problems like using an MRI model for CT images~\cite{ouyang2019data}, we argue that in real practice the minor domain shift from scanners and imaging parameters are the most common problems, and we propose a convenient and novel way to alleviate it. Future work will be improving the adaptation results by incorporating self-supervised methods~\cite{carlucci2019domain} and improved auto-encoders~\cite{gong2019memorizing}.

\section{Acknowledgments}
This work is supported by NIH grants R01-EY024655 (PI: J.L. Prince), R01-NS082347 (PI: P.A. Calabresi) and in part by the Intramural research Program of the NIH, National Institute on Aging.
\bibliographystyle{splncs04}
\bibliography{ml,oct}

\end{document}